\documentclass[mlmain,onecolumn,pmlr]{jmlr}

\usepackage{longtable}

\usepackage{booktabs}
\usepackage[load-configurations=version-1]{siunitx} 

\theorembodyfont{\upshape}
\theoremheaderfont{\scshape}
\theorempostheader{:}
\theoremsep{\newline}

\usepackage{array} 
\usepackage{multirow} 
\usepackage{listings}

\definecolor{deepblue}{rgb}{0.29411765 0.45882353 0.61960784}
\definecolor{deepred}{rgb}{0.74509804 0.21176471 0.23921569}
\definecolor{deepgreen}{rgb}{0,0.5,0}
\definecolor{deeppurple}{rgb}{0.52941176 0.32941176 0.56470588}
\definecolor{codegray}{rgb}{0.5,0.5,0.5}
\definecolor{backcolour}{rgb}{0.95,0.95,0.92}

\lstdefinestyle{mystyle}{
    backgroundcolor=\color{backcolour},
    commentstyle=\color{deeppurple},
    keywordstyle=\color{deepred},
    numberstyle=\tiny\color{codegray},
    stringstyle=\color{deepblue},
    basicstyle=\ttfamily\small\linespread{1.1},
    breakatwhitespace=false,
    breaklines=true,
    captionpos=b,
    keepspaces=true,
    numbers=left,
    numbersep=8pt,
    showspaces=false,
    showstringspaces=false,
    showtabs=false,
    frame=single,
    framerule=0.4pt,
    rulecolor=\color{codegray},
    tabsize=4,
    aboveskip=1.5ex,
    belowskip=1.5ex,
    xleftmargin=15pt,
    xrightmargin=15pt,
    extendedchars=true,
    columns=flexible,
    linewidth=\textwidth,
    literate={-}{-}1
}

\usepackage{textcomp}

\usepackage[capitalize,noabbrev]{cleveref}

\usepackage{todonotes}

\newcolumntype{T}[1]{>{\centering\arraybackslash}m{#1}}

\jmlrvolume{}
\firstpageno{1}

\jmlryear{2024}
\jmlrworkshop{Symmetry and Geometry in Neural Representations}

\title[A Cosmic-Scale Benchmark for Symmetry-Preserving Data Processing]{A Cosmic-Scale Benchmark for Symmetry-Preserving \titlebreak Data Processing}

  \author{\Name{Julia Balla} \Email{jballa@mit.edu}\\
  \Name{Siddharth Mishra-Sharma}\footnote{Currently at Anthropic; work performed while at MIT/IAIFI.}\Email{smsharma@mit.edu}\\
  \Name{Carolina Cuesta-Lazaro} \Email{cuestalz@mit.edu}\\
  \Name{Tommi Jaakkola} \Email{tommi@csail.mit.edu}\\
  \Name{Tess Smidt} \Email{tsmidt@mit.edu}\\
  \addr Massachusetts Institute of Technology, IAIFI
  }

\begin{document}

\maketitle

\begin{abstract}
Efficiently processing structured point cloud data while preserving multiscale information is a key challenge across domains, from graphics to atomistic modeling. Using a curated dataset of simulated galaxy positions and properties, represented as point clouds, we benchmark the ability of graph neural networks to simultaneously capture local clustering environments and long-range correlations. Given the homogeneous and isotropic nature of the Universe, the data exhibits a high degree of symmetry. We therefore focus on evaluating the performance of Euclidean symmetry-preserving ($E(3)$-equivariant) graph neural networks, showing that they can outperform non-equivariant counterparts and domain-specific information extraction techniques in downstream performance as well as simulation-efficiency. However, we find that current architectures fail to capture information from long-range correlations as effectively as domain-specific baselines, motivating future work on architectures better suited for extracting long-range information.
\end{abstract}

\section{Introduction}

Point clouds are discrete elements in a coordinate system, defined by spatial coordinates and optional attributes like velocities. Their unordered structure often reveals intricate geometric patterns that span multiple scales, from local neighborhoods to global distributions. Extracting meaningful insights from such data motivates the development of new machine learning algorithms that are capable of capturing and exploiting these multiscale features. Scientific datasets provide an ideal setting for stress-testing these algorithms, as they often exhibit highly structured, yet low-dimensional latent representations despite their complex, high-dimensional observational data. Cosmology -- the study of the Universe's origin, structure, and evolution -- is a prime example of this since the laws driving the Universe's origin, structure, and evolution are amenable to relatively `simple' descriptions, allowing scientific data to be characterized using low-dimensional summaries. 

A particular type of flagship observation in cosmology is \emph{galaxy clustering}, where the positions and associated properties of galaxies are measured by cosmological surveys. Different spatial distributions of these galaxies can provide insights into the underlying structure of the Universe, targeting questions like the distribution and nature of dark matter and the expansion history of the Universe. Reliably extracting information from these surveys is a challenging task that is becoming increasingly timely as next-generation cosmological surveys like the Dark Energy Spectroscopic Instrument (DESI) \citep{levi2019dark} are set to deliver petabytes of data in the form of point clouds consisting of over $\mathcal O(10^6)$ points, necessitating the development of new data-processing and compression algorithms based on simulations to fully realize their scientific goals. Furthermore, the simulations are extremely computationally expensive, making simulation-efficiency a key consideration in the development of new algorithms for this domain.

Cosmological datasets and associated tasks exhibit several distinguishing features that make them a valuable benchmark for stress-testing and developing novel machine learning algorithms, like graph neural networks, for processing point cloud data. Examples include:
\begin{itemize}
    \item \textbf{Point cloud cardinality}: The datasets under consideration are larger than those commonly encountered in other scientific domains where graph processing is used, such as the study of atomistic systems, with $\mathcal O(10-100)$ points. This presents unique challenges when it comes to scalability and processing information across point clouds.
    \item \textbf{Information across scales}: Gravitational forces cause matter to cluster, leading to strong small-scale correlations. On the other hand, growth of structures that were initially in causal contact but are spatially separated at present times induces long-range correlations. This multiscale nature necessitates the use of algorithms that can capture both local and global information.
    \item \textbf{Symmetry structure}: The Universe is homogeneous and isotropic -- its properties are spatially uniform. This implies that the distribution of galaxies and other cosmic structures should exhibit Euclidean symmetry (i.e., invariance to translations, rotations, and reflections).
\end{itemize}

Although the data is not inherently graph-structured (rather, it is a set of points without specified ordering or interconnectedness), graph neural networks (GNNs) provide an efficient way to process this data, and we focus on benchmarking these here.

\paragraph{Primary contributions}

Our primary contributions are as follows. First, we curate a dataset from existing cosmological $N$-body simulations tailored towards benchmarking point cloud processing algorithms and introduce an easy-to-use interface for accessing the dataset, without the need for specialized domain expertise. Simultaneously, we introduce a Jax-based code repository, \texttt{eqnn-jax}, implementing common equivariant neural networks customized for processing point cloud data with associated physical features and using them for inference tasks. Second, we systematically study the effect of various architectural choices on the performance of graph neural networks on downstream tasks using this dataset, with a particular focus on those that incorporate the relevant symmetries. Finally, we compare the performance of these models against traditional domain-informed summary statistics, with the goal of assessing the potential of machine learning methods to automate and improve upon existing data analysis techniques in cosmology. 

\section{Background and Related Work}

\paragraph{Cosmology with point clouds} 
Previous works have explored point cloud-based approaches to extracting information from galaxy distributions. \citet{makinen2022cosmic} used GNNs to extract cosmological parameters from galaxy positions, focusing on a specific architecture and relatively small point clouds with $\mathcal{O}(100)$ points. \citet{villanueva2022learning} again considered $\mathcal{O}(1000)$ galaxy points with limited architectural variation. \citet{anagnostidis2022cosmology} used a non-graph-based approach with PointNet++~\citep{qi2017pointnet,qian2022pointnext}. Our goal is to systematically study the performance and efficiency of various GNNs, focusing for the first time on symmetry-preserving architectures for larger datasets with $\mathcal{O}(10^4)$ points. While conventional 3D computer vision datasets \citep{chang2015shapenet,Zhirong15CVPR} often have high cardinality, they emphasize structured shapes, whereas our dataset requires models to capture multi-scale structures across entire point clouds.

\paragraph{Long-range graph benchmarks} Benchmarking GNNs is an active research area, with recent studies highlighting the difficulty of capturing long-range correlations due to oversmoothing and oversquashing effects \citep{dwivedi2022long,tonshoff2023did}. \citet{dwivedi2022long} introduced the Long Range Graph Benchmark (LRGB), showing that graph transformers significantly outperform vanilla message-passing GNNs on tasks requiring long-range correlations. However, unlike the LRGB datasets, which are limited to 500 nodes, our benchmark requires processing long-range information across a larger set of points.

\paragraph{Equivariant graph neural networks} Neural networks that incorporate physically-informed inductive biases, e.g. symmetries \citep{geiger2022e3nn}, have been demonstrated to be effective in a variety of domains, from particle physics to materials science \citep{batatia2023foundation,batzner20223,batatia2022mace,jumper2021highly}. 
In equivariant neural networks, described in more detail below, for a transformation $T$ of the input $x$, the output $f(x)$ transforms correspondingly: $f(T(x)) = T'(f(x))$, where $T'$ is a transformation related to $T$. These networks can be more simulation-efficient, a requirement in many data-poor settings where data may be expensive to generate, and can also guarantee the validity of output configurations when these contain strict symmetry requirements. 
Equivariant neural networks have also been shown to scale more favorably with training set size \citep{batzner20223}. On the other hand, some recent applications question the need to incorporate symmetries for downstream performance \citep{abramson2024accurate,wang2023generating}. Equivariant neural networks have previously been benchmarked in other physical sciences domains, e.g. fluid mechanics \citep{toshev2024lagrangebench}. Our benchmark aims to stress-test existing symmetry-sensitive architectures in a novel, challenging setting.

\section{Dataset and Benchmark Tasks} \label{sec:dataset}

\subsection{Galaxy Clustering Dataset}

\paragraph{Simulation and processing details} Our dataset is derived from Sobol sequence set of the \emph{Quijote} suite of $N$-body simulations \citep{villaescusa2020quijote}.
\footnote{The \emph{Quijote} simulation dataset and, in particular, the Big Sobol Sequence (BSQ) is available at\\ \url{https://quijote-simulations.readthedocs.io/en/latest/bsq.html}.} These simulations model the evolution of $512^3$ dark matter particles in a periodic comoving volume of $\sim 1$ Gigaparsec. Each simulation varies the cosmological parameters and the random phases of the initial conditions simultaneously, providing a diverse range of possible universes to learn from and test on. The simulations employ periodic boundary conditions, which ensure that the Universe is modeled as statistically homogeneous and isotropic on large scales. This treatment allows for the accurate modeling of the large-scale structure without introducing artificial boundary effects. 

Dark matter halos, which are gravitationally bound structures that host galaxies, are identified in the simulations using a halo-finding algorithm. We do not distinguish between dark matter halos and galaxies in this work, neglecting the details of the connection between galaxies and dark matter, which are not significant for benchmarking purposes.  These simulations are computationally expensive to run, with over 35 million CPU hours required to generate 44,100 simulations for the initial suite \citep{villaescusa2020quijote}. This computational cost highlights the need for simulation-efficient methods to extract information from the resulting datasets, which is one of the primary motivations for our work. Further details of the simulation are given in App. \ref{app:dataset}.

\begin{figure*}[!tbp]
\centering
\begin{tabular}{@{}c@{}}
\includegraphics[width=\linewidth]{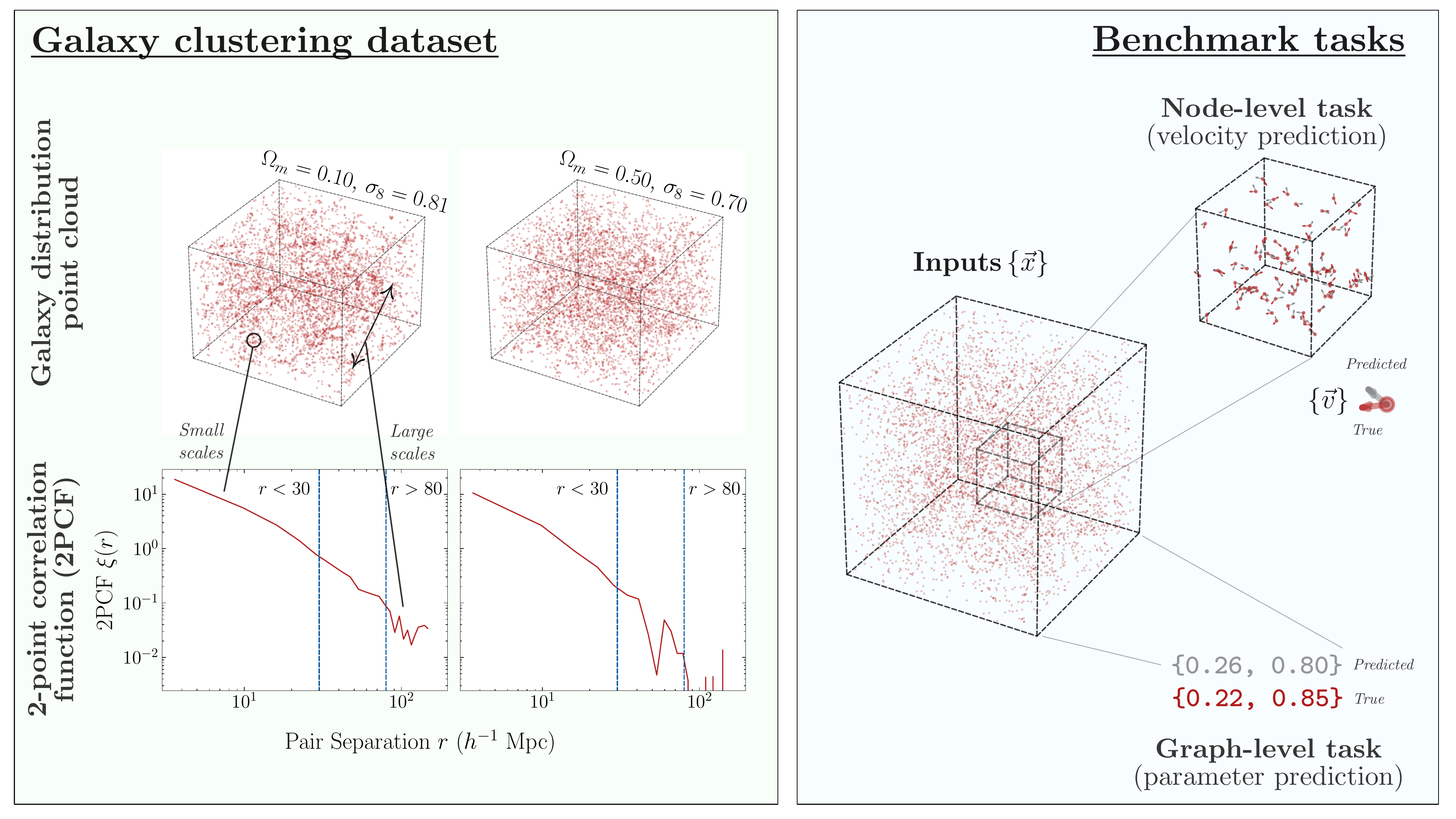}\\ 
\end{tabular}
\caption{
    (Left)  Exemplary point clouds from the training set and their corresponding 2-point correlation functions. (Right) An illustration of the benchmark tasks.
    }
    \label{fig:dataset}
\end{figure*}

\paragraph{Dataset description} The resulting dark matter halo coordinates are represented as a 3D point cloud. We select the 5000 most massive halos by halo mass in each simulation to construct our dataset. This selection by number of objects, rather than by a minimum halo mass threshold, mimics the constraints of observational data where masses are unknown. The final dataset consists of point clouds of shape $\mathbb{R}^{5000 \times 3}$, with each point representing the 3D position of a dark matter halo. In addition to the halo coordinates, the data also contains other attributes, namely halo mass, as well the velocity and angular momentum vectors for each halo. In the current benchmarks, we utilize the velocity attributes for a subset of our experiments, resulting in point clouds of shape $\mathbb{R}^{5000 \times 6}$. A total of 12,384 simulations are available, from which a subset is split into a training set of size 2048 and validation and test sets of size 512 for benchmarking. We show how the different architectures scale with training set size in Section~\ref{sec:scaling}. Examples of point cloud samples are visualized in Fig.~\ref{fig:dataset}.

\paragraph{Two-point correlation function (2PCF)} A key statistical measure of the large-scale structure is the isotropic two-point correlation function (2PCF), which quantifies the excess probability of finding pairs of galaxies at a given separation compared to a random distribution. The 2PCF is an efficient summary statistic because it encodes information about the clustering of galaxies at different scales, can be accurately measured from observational data, and is well-understood theoretically, making it a powerful tool for constraining cosmological models. In this work, we use the 2PCF as a baseline to compare against our models, highlighting their potential to extract more information from the point cloud data than traditional summary statistics. We include the 2PCF as part of the dataset, computed as a 24-dimensional vector where each dimension represents a logarithmically spaced bin in radial distance. 2PCFs corresponding to example point clouds are visualized in Fig.~\ref{fig:dataset}.

\subsection{Data Access}

To facilitate easy access to the dataset used in this work, we provide a high-level Python interface (described in App.  \ref{app:data_access}) for loading and preprocessing the point cloud data derived from the processed simulation data. The raw data is stored in the TFRecord format, allowing for efficient storage and retrieval. The dataset and code for all experiments are available at \url{https://github.com/smsharma/eqnn-jax}.

\subsection{Benchmark Tasks}

We consider two benchmark tasks to evaluate the performance of our models: a graph-level and a node-level prediction task. These tasks are designed to test the ability of the models to learn global properties of the point cloud as well as leverage correlations in the local environment, respectively. We note that we don't model the dynamics of the system, which is otherwise a common task for benchmarking (equivariant) GNNs \citep{toshev2024lagrangebench}.

\paragraph{Graph-level prediction} The graph-level prediction task is a regression problem where the goal is to infer two key cosmological parameters from an input point cloud. Specifically, given a point cloud $\mathbf{X} \in \mathbb{R}^{5000 \times 3}$, representing 5000 galaxy positions, the task is to predict the values of the matter density ($\Omega_m$) and the root-mean-square matter fluctuation averaged over a sphere of radius $\sim 8$ Mpc ($\sigma_8$), the degree of inhomogeneity in the matter distribution on these scales; $f: \mathbb{R}^{5000 \times 3} \rightarrow \mathbb{R}^{2}$. These parameters are fundamental to describing the structure of the Universe and are primary targets of current and upcoming cosmological surveys. $\Omega_m$ tends to depend sensitively on the nature of long-range correlations, while $\sigma_8$ captures information about local correlations. We train the models using the mean squared error (MSE) loss between the predicted output and true target parameters.

\paragraph{Node-level prediction} The node-level prediction task is a regression problem designed to test the ability of the models to capture local dependencies within the point cloud, while outputting a more manifestly ``geometric'' quantity -- a velocity vector for each node. 
The input is again a point cloud $\mathbf{X} \in \mathbb{R}^{5000 \times 3}$, where each row corresponds to a galaxy. 
The output is a tensor $\mathbf{Y} \in \mathbb{R}^{5000 \times 3}$ representing the predicted velocity components for all points; $f: \mathbb{R}^{5000 \times 3} \rightarrow \mathbb{R}^{5000 \times 3}$. We train the models using the MSE loss on the predicted velocities. Note that this is a very challenging task when only considering the heaviest $5000$ galaxies.

\paragraph{Relevance to model performance interpretation} The different tasks and target parameters probe different abilities of GNN information extraction, highlighting the usefulness of scientific datasets as benchmarks. The parameter $\Omega_m$ tends to depend sensitively on the nature of long-range correlations, while $\sigma_8$ captures information about small-scale gravitational clustering. The ability of a model to perform well at inferring these target parameters is hence indicative of their ability to leverage long- and short-range correlations, respectively. Inferring velocities (the node task) on the other hand requires leveraging the structure of the local environment around a point, without averaging over the point cloud, but also learning to capture long range correlations that introduce large scale motions. 

\section{Architectures and Baselines}
The graph neural networks we utilize follow the general message-passing framework based on \citet{battaglia2018relational}. A local $k$-nearest neighbors graph is constructed using the Euclidean distance between coordinates as the distance metric, accounting for periodic boundary conditions across the box edges. The graph is represented by node features $\vec{x}_i$ (positions) and edge features $\vec{e}_{ij}$ (relative distances). We project relative distances onto a basis of radial Bessel functions with a radial cutoff of 0.6 on the $Z$-scored positions, which was found to be crucial for downstream performance in the graph-level prediction task to predict $\sigma_8$, the parameter most affected by short range correlations. 

Specific implementations differ in the choice of edge/node update functions, and features used in message passing. Graph Neural Network (GNN) closely follows the general framework outlined above, using MLPs for the edge and node update functions $\phi_e^l$ and $\phi_x^l$. $E(n)$ Equivariant Graph Neural Network (EGNN) \citep{satorras2021n} designs the edge and node updates such that the message-passing operation is equivariant to $E(n)$ transformations. Steerable $E(3)$ Equivariant Graph Neural Network (SEGNN) \citep{brandstetter2021geometric} utilizes steerable feature representations, allowing the node and edge features to be covariant geometric tensors of arbitrary order (e.g. vectors, higher-order tensors) rather than just invariant scalars. Neural Equivariant Interatomic Potential (NequIP) \citep{batzner20223} also uses steerable feature representations, constructing equivariant message passing layers using Clebsch-Gordan tensor products and spherical harmonics. SEGNN and NequIP use a hyperparameter $\ell_\mathrm{max}$ to control the maximum degree of the spherical harmonics used in the steerable feature representations, with higher values allowing the models to capture more complex angular dependencies at computational cost.

We compare the above MPNN-based methods with PointNet++ \citep{qian2022pointnext}, which processes a set of points sampled in a metric space in a hierarchical fashion.  See App. \ref{app:architectures} for a more detailed description of each architecture.

\section{Experiments and Discussion} \label{sec:experiments}

\subsection{Training details}

Models were trained for 5000 steps with 5-fold cross validation using the AdamW optimizer \citep{loshchilov2017decoupled} and a cosine decay schedule. The checkpoint corresponding to the lowest validation loss is used for evaluation. Training was performed on 4 NVIDIA A100 GPUs and took approximately {14--44 minutes per training run, depending on the specific architecture used. For instance, the SEGNN with $\ell_\mathrm{max}=1$ runs approximately 4.42 iterations per second (it/s) on the graph task, compared to 2.40it/s with $\ell_\mathrm{max}=2$. As an equivariant model, NequIP is much leaner at 7.06 it/s. App. \ref{app:hyperparams} contains the model hyperparameters used in all of the following experiments.

\subsection{Graph- and node-level prediction}

\begin{figure*}[!tbp]
\centering
    \includegraphics[width=0.49\textwidth]{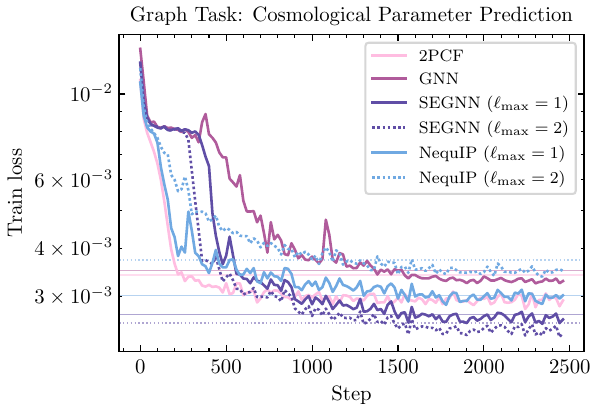}
    \includegraphics[width=0.4827\textwidth]{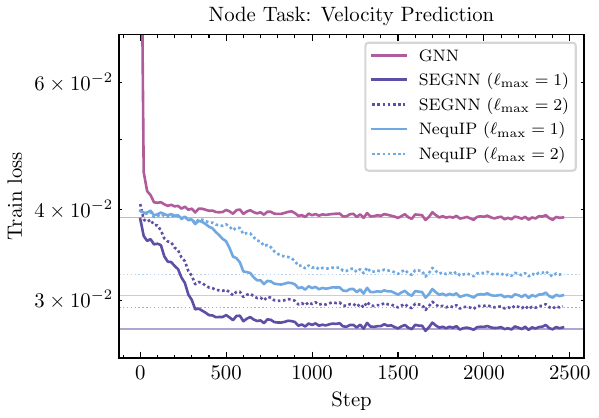}
    \caption{Training losses over the course of training for various models considered, for the graph-level prediction task (left) and the node-level prediction task (right). The final test loss is shown as a horizontal dashed line.}
    \label{fig:training}
\end{figure*}

Figure \ref{fig:training} shows the losses over the course of training for the graph-level prediction task and the node-level prediction task. The final test losses are shown as horizontal lines. Table \ref{tab:best_models} compares the test-set performance of different models on the two tasks, along with the number of parameters for the best-performing model. Three sets of results are shown, separated by horizontal lines -- those where \emph{(1)} the input point cloud consists of just position coordinates, \emph{(2)} where the point clouds additionally include a velocity vector for each galaxy, and \emph{(3)} where the two-point correlation function is used as an additional global context input (see Sec. \ref{sec:tpcf_global} below).

We see that the equivariant models (SEGNN and NequIP) outperform the non-equivariant models (GNN and PointNet++) for both tasks. Higher spherical harmonic orders $\ell_\mathrm{max}$ provide benefit for the velocity prediction task for SEGNN, but not in the other cases. While equivariant models beat the non-equivariant ones in graph-level parameter prediction, the domain-informed 2PCF summary shows superior performance in extracting $\Omega_m$, which requires capturing long-range correlations; this indicates that there is yet information left on the table when using message-passing GNNs, in particular when it comes to long-range correlations. Equivariant models also show faster convergence than the non-equivariant counterpart. 

When node velocities are provided as an additional input feature, both the GNN and SEGNN show significant improvements on graph-level tasks. However, when velocities are used solely as node features in the SEGNN, its performance is worse than the GNN, likely due to a more complex loss landscape that requires additional hyperparameter tuning \citep{flinth2023optimization}. The key gain for the SEGNN comes from using velocities as steerable attributes, allowing it to outperform the GNN when $\ell_\mathrm{max} = 2$. Additionally, given the demonstrated effectiveness of transformer models in capturing long-range dependencies through attention mechanisms \citep{dwivedi2022long}, we include an ablation study of local and global attention-based aggregation mechanisms in GNN layers in App. \ref{app:gnn_attn}.

\begin{table*}[t]
\small
\centering
\setlength{\tabcolsep}{4pt}
\renewcommand{\arraystretch}{1.1}
\begin{tabular*}{\textwidth}{@{\extracolsep{\fill}} c l T{1.75cm} T{1.75cm} T{1cm} T{1.75cm} T{1cm}@{}}
\toprule
\multicolumn{1}{c}{\raisebox{1ex}{\multirow{3}{*}{\rotatebox[origin=c]{90}{Inputs}}}} & \multicolumn{1}{c}{\multirow{3}{*}{Model}} & \multicolumn{3}{c}{Graph task} & \multicolumn{2}{c}{Node task} \\ 
\cmidrule(lr){3-5} \cmidrule(lr){6-7}
& & \multicolumn{1}{c}{$\Omega_m$} & \multicolumn{1}{c}{$\sigma_8$} & \multicolumn{1}{c}{Params.} & \multicolumn{1}{c}{$\vec{v}$} & \multicolumn{1}{c}{Params.} \\ 
\midrule
\multirow{7}{*}{\rotatebox[origin=c]{90}{$\vec x$}} & 2PCF & $\mathbf{2.03 \pm 0.02}$ & $4.66 \pm 0.06$ & 56k & {$-$} & {$-$} \\ 
& GNN & $2.77 \pm 0.41$  & $4.84 \pm 2.90$ & 1441k & $2.94 \pm 0.03$ & 463k \\ 
& EGNN & $13.33 \pm 0.00$ & $13.37 \pm 0.00$ & 342k & {$-$} & {$-$} \\ 
& NequIP ($\ell_\mathrm{max} = 1$) & $2.88 \pm 0.15$ & $5.05 \pm 1.08$ & 439k & $2.28
\pm
0.00$ & 154k \\ 
& NequIP ($\ell_\mathrm{max} = 2$) & $3.07 \pm 0.18$ & $4.80 \pm 0.49$ & 450k & $2.44 \pm 0.00$ & 163k \\ 
& SEGNN ($\ell_\mathrm{max} = 1$) & $2.31 \pm 0.03$ & 
$\mathbf{2.34 \pm 0.08}$ & 1015k &  $2.06 \pm 0.00$ & 280k \\ 
& SEGNN ($\ell_\mathrm{max} = 2$) & $2.37 \pm 0.06$ & $2.36 \pm 0.22$ & 1458k & $\mathbf{2.04 \pm 0.00}$ & 401k \\ 
& PointNet++ & $2.87 \pm 0.07$ & $9.00 \pm 3.94$ & 1354k & $2.92 \pm 0.00 $ & 463k\\
\midrule
\multirow{3}{*}{\rotatebox[origin=c]{90}{$\vec x, \vec v$}} & GNN &  $1.10 \pm 0.02$ &  $1.96 \pm 0.04$ & 702k & {$-$} & {$-$} \\
& SEGNN ($\ell_\mathrm{max} = 1$) & $1.16 \pm 0.02$ & $1.65 \pm 0.02$ & 654k & {$-$} & {$-$} \\
& SEGNN ($\ell_\mathrm{max} = 2$) & $1.13 \pm 0.03$ & $1.76 \pm 0.07$ & 876k & {$-$} & {$-$} \\
& SEGNN ($\ell_\mathrm{max} = 1$, steerable $\vec v$) & $0.99 \pm 0.03$ & $1.86 \pm 0.04$ & 654k & {$-$} & {$-$} \\
& SEGNN ($\ell_\mathrm{max} = 2$, steerable $\vec v$) & $\mathbf{0.84 \pm 0.01}$ & $\mathbf{1.42 \pm 0.02}$ & 876k & {$-$} & {$-$} \\
\midrule
\multirow{3}{*}{\rotatebox[origin=c]{90}{\footnotesize $\vec x, \mathrm{2PCF}$}} & SEGNN ($\ell_\mathrm{max} = 2$) + 2PCF &   
$\mathbf{1.66 \pm 0.01}$  & $2.38 \pm 0.07$ & 1543k & {$-$} & {$-$} \\
& SEGNN ($\ell_\mathrm{max} = 2$) + 2PCF\textsubscript{small} & $2.27 \pm 0.01$ & $2.40 \pm 0.04$ & 1504k & {$-$} & {$-$} \\
& SEGNN ($\ell_\mathrm{max} = 2$) + 2PCF\textsubscript{large} & $1.73 \pm 0.04$ & $\mathbf{2.26 \pm 0.09}$ & 1512k & {$-$} & {$-$} \\
\bottomrule
\end{tabular*}
\caption{Comparison of different models on the graph- and node-level tasks. All mean-squared error values of $\Omega_m$ and $\sigma_8$ are in units of $10^{-3}$.  Variance is reported up to the second decimal place. The best results for each section are shown in \textbf{bold}.}
\label{tab:best_models}
\end{table*}

\subsection{Two-point correlation function as global information}
\label{sec:tpcf_global}

To gain insight into the information captured by the 2PCF that is not captured by any of the graph neural network models (as evidenced by their worse performance when predicting $\Omega_m$), we evaluate the effects of adding the 2PCF as a global input feature. In particular, we select the best-performing model on both tasks: SEGNN with $\ell_\mathrm{max} = 2$. After the final message-passing layer, the pooled graph-wise representation is concatenated with the 2PCF before being fed into the readout MLP. We also compare the effects of using only subsections of the 2PCF that correspond to small-scale ($r<30$ $h^{-1}$ Mpc) and large-scale information ($r>80$ $h^{-1}$ Mpc). These are shown in the last set of runs in Tab. \ref{tab:best_models}. The SEGNN matches the performance of the 2PCF baseline on $\Omega_m$ prediction when it is equipped with the large-scale 2PCF components. Moreover, it drastically outperforms the baseline when the whole 2PCF vector is used, suggesting that there is crucial information present in the mid-range scales as well. For the task of $\sigma_8$ prediction, which relies on capturing local correlations, the small-scale 2PCF information helps as much as the full 2PCF vector.

\subsection{Scaling with dataset size}
\label{sec:scaling}
Figure \ref{fig:scaling} shows the test-set performance of various models as a function of the number of samples in the training dataset, for the graph-level task. The equivariant SEGNN models, in particular, show better performance at all training sample sizes, while being more simulation-efficient. Higher spherical harmonic order help marginally when using SEGNN, but not NequIP. We see that, in addition to performing better in the data-poor regime, the relative advantage of equivariant models is maintained in the asymptotic regime, up to the number of simulations available ($\sim 10^4$).

\begin{figure*}[!tbp]
\centering
    \includegraphics[width=0.65\textwidth]{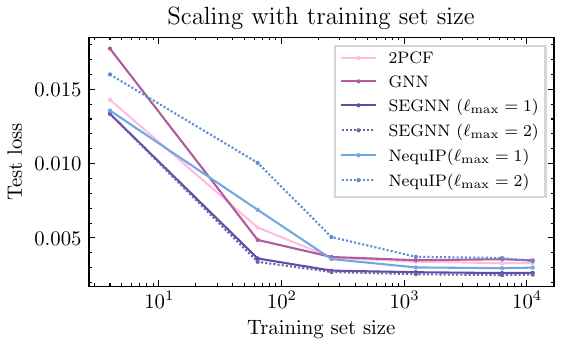}
    \caption{Test loss scaling on the graph task as a function of training set size across models.}
    \label{fig:scaling}
\end{figure*}

\section{Conclusion}
\label{sec:conclusions}

We investigated the ability of graph neural network architectures, with a focus on symmetry-preserving variants, to extract short- and long-range information from point cloud data using cosmology data. The benchmark dataset consists of positions of simulated galaxies, whose spatial distribution is informative of the underlying cosmological model. We showed that both graph-level (i.e., inferring cosmological parameters) and node-level (i.e., inferring the 3D velocity of each galaxy) prediction tasks can benefit from the use of equivariant models, which were also found to be more simulation-efficient. This is particularly relevant for the domain under study, where producing new simulations is compute-intensive. Equivariant models can therefore enable practitioners to do more with available simulations. 

However, we also found that the domain-specific two-point correlation function (2PCF) summary statistic outperformed the GNNs in inferring the cosmological parameter $\Omega_m$, which is sensitive to long-range correlations. By incorporating the 2PCF as a global context vector in the SEGNN architecture, we were able to significantly improve its performance, surpassing that of the 2PCF alone, demonstrating the potential of combining domain knowledge with machine learning models. Message-passing GNNs are known to struggle with long-range correlations \citep{Li_Han_Wu_2018, alon2021on}, and this dataset provides a benchmark to probe their ability to effectively leverage these. The present benchmark would be a good target for methods that aim to mitigate issues associated with long-range information preservation \citep{jain2021representing, topping_understanding_2022, karhadkar2023fosr}.

\paragraph{Limitations and future work} This work does not include many recent architectures, such as GATr \citep{brehmer2023geometric}, MACE \citep{batatia2022mace}, or P$\Theta$NITA \citep{bekkers2023fast} as our goal was not to achieve the best possible performance. Instead, we focused on isolating the effects of key architectural components, such as invariance and equivariance, on tasks involving local and long-range correlations. By selecting a few representative models, we aimed for a focused analysis rather than testing a wide range of architectures. Future work can leverage our dataset to benchmark more advanced models. The equivariant architectures we studied were either general-purpose in nature (e.g., SEGNN) or designed for specific domains (e.g., NequIP for atomistic systems). Our results motivate the development of specialized architectures tailored to cosmology data, which would be sensitive to the local gravitational clustering environment as well as the long-range correlations in galaxy fields. 

The benchmark presented is aimed at machine learning researchers as well as domain scientists in cosmology. From the perspective of machine learning, the benchmark provides a challenging and realistic test bed for the development of novel equivariant graph neural network architectures that can handle long-range correlations and scale to large datasets. From the perspective of cosmology, our results demonstrate the potential of machine learning techniques to extract more information from expensive simulations and to automate and improve upon traditional data analysis pipelines. We hope that this work will spur bidirectional collaboration between the two communities.

\acks{This work is supported by the National Science Foundation under Cooperative Agreement PHY-2019786 (The NSF AI Institute for Artificial Intelligence and Fundamental Interactions, \url{http://iaifi.org/}). This material is based upon work supported by the U.S. Department of Energy, Office of Science, Office of High Energy Physics of U.S. Department of Energy under grant Contract Number DE-SC0012567. This work was performed in part at the Aspen Center for Physics, which is supported by NSF grants PHY-2210452. The computations in this paper were run on the FASRC Cannon cluster supported by the FAS Division of Science Research Computing Group at Harvard University. This work is also supported by the Department of Defense (DoD) through the National Defense Science \& Engineering Graduate (NDSEG) Fellowship Program.}

\bibliography{references}

\begin{thebibliography}{32}
\providecommand{\natexlab}[1]{#1}
\providecommand{\url}[1]{\texttt{#1}}
\expandafter\ifx\csname urlstyle\endcsname\relax
  \providecommand{\doi}[1]{doi: #1}\else
  \providecommand{\doi}{doi: \begingroup \urlstyle{rm}\Url}\fi

\bibitem[Abramson et~al.(2024)Abramson, Adler, Dunger, Evans, Green, Pritzel, Ronneberger, Willmore, Ballard, Bambrick, et~al.]{abramson2024accurate}
Josh Abramson, Jonas Adler, Jack Dunger, Richard Evans, Tim Green, Alexander Pritzel, Olaf Ronneberger, Lindsay Willmore, Andrew~J Ballard, Joshua Bambrick, et~al.
\newblock Accurate structure prediction of biomolecular interactions with alphafold 3.
\newblock \emph{Nature}, pages 1--3, 2024.

\bibitem[Alon and Yahav(2021)]{alon2021on}
Uri Alon and Eran Yahav.
\newblock On the bottleneck of graph neural networks and its practical implications.
\newblock In \emph{International Conference on Learning Representations}, 2021.

\bibitem[Anagnostidis et~al.(2022)Anagnostidis, Thomsen, Kacprzak, Tr{\"o}ster, Biggio, Refregier, and Hofmann]{anagnostidis2022cosmology}
Sotiris Anagnostidis, Arne Thomsen, Tomasz Kacprzak, Tilman Tr{\"o}ster, Luca Biggio, Alexandre Refregier, and Thomas Hofmann.
\newblock Cosmology from galaxy redshift surveys with pointnet.
\newblock \emph{arXiv preprint arXiv:2211.12346}, 2022.

\bibitem[Batatia et~al.(2022)Batatia, Kovacs, Simm, Ortner, and Cs{\'a}nyi]{batatia2022mace}
Ilyes Batatia, David~P Kovacs, Gregor Simm, Christoph Ortner, and G{\'a}bor Cs{\'a}nyi.
\newblock Mace: Higher order equivariant message passing neural networks for fast and accurate force fields.
\newblock \emph{Advances in Neural Information Processing Systems}, 35:\penalty0 11423--11436, 2022.

\bibitem[Batatia et~al.(2023)Batatia, Benner, Chiang, Elena, Kov{\'a}cs, Riebesell, Advincula, Asta, Baldwin, Bernstein, et~al.]{batatia2023foundation}
Ilyes Batatia, Philipp Benner, Yuan Chiang, Alin~M Elena, D{\'a}vid~P Kov{\'a}cs, Janosh Riebesell, Xavier~R Advincula, Mark Asta, William~J Baldwin, Noam Bernstein, et~al.
\newblock A foundation model for atomistic materials chemistry.
\newblock \emph{arXiv preprint arXiv:2401.00096}, 2023.

\bibitem[Battaglia et~al.(2018)Battaglia, Hamrick, Bapst, Sanchez-Gonzalez, Zambaldi, Malinowski, Tacchetti, Raposo, Santoro, Faulkner, et~al.]{battaglia2018relational}
Peter~W Battaglia, Jessica~B Hamrick, Victor Bapst, Alvaro Sanchez-Gonzalez, Vinicius Zambaldi, Mateusz Malinowski, Andrea Tacchetti, David Raposo, Adam Santoro, Ryan Faulkner, et~al.
\newblock Relational inductive biases, deep learning, and graph networks.
\newblock \emph{arXiv preprint arXiv:1806.01261}, 2018.

\bibitem[Batzner et~al.(2022)Batzner, Musaelian, Sun, Geiger, Mailoa, Kornbluth, Molinari, Smidt, and Kozinsky]{batzner20223}
Simon Batzner, Albert Musaelian, Lixin Sun, Mario Geiger, Jonathan~P Mailoa, Mordechai Kornbluth, Nicola Molinari, Tess~E Smidt, and Boris Kozinsky.
\newblock E (3)-equivariant graph neural networks for data-efficient and accurate interatomic potentials.
\newblock \emph{Nature communications}, 13\penalty0 (1):\penalty0 2453, 2022.

\bibitem[Behroozi et~al.(2012)Behroozi, Wechsler, and Wu]{Behroozi_2012}
Peter~S. Behroozi, Risa~H. Wechsler, and Hao-Yi Wu.
\newblock The rockstar phase-space temporal halo finder and the velocity offsets of cluster cores.
\newblock \emph{The Astrophysical Journal}, 762\penalty0 (2):\penalty0 109, December 2012.
\newblock ISSN 1538-4357.

\bibitem[Bekkers et~al.(2023)Bekkers, Vadgama, Hesselink, van~der Linden, and Romero]{bekkers2023fast}
Erik~J Bekkers, Sharvaree Vadgama, Rob~D Hesselink, Putri~A van~der Linden, and David~W Romero.
\newblock Fast, expressive se $(n) $ equivariant networks through weight-sharing in position-orientation space.
\newblock \emph{arXiv preprint arXiv:2310.02970}, 2023.

\bibitem[Brandstetter et~al.(2021)Brandstetter, Hesselink, van~der Pol, Bekkers, and Welling]{brandstetter2021geometric}
Johannes Brandstetter, Rob Hesselink, Elise van~der Pol, Erik~J Bekkers, and Max Welling.
\newblock Geometric and physical quantities improve e (3) equivariant message passing.
\newblock \emph{arXiv preprint arXiv:2110.02905}, 2021.

\bibitem[Brehmer et~al.(2023)Brehmer, de~Haan, Behrends, and Cohen]{brehmer2023geometric}
Johann Brehmer, Pim de~Haan, S{\"o}nke Behrends, and Taco Cohen.
\newblock Geometric algebra transformer.
\newblock In \emph{Advances in Neural Information Processing Systems}, volume~37, 2023.

\bibitem[Chang et~al.(2015)Chang, Funkhouser, Guibas, Hanrahan, Huang, Li, Savarese, Savva, Song, Su, et~al.]{chang2015shapenet}
Angel~X Chang, Thomas Funkhouser, Leonidas Guibas, Pat Hanrahan, Qixing Huang, Zimo Li, Silvio Savarese, Manolis Savva, Shuran Song, Hao Su, et~al.
\newblock Shapenet: An information-rich 3d model repository.
\newblock \emph{arXiv preprint arXiv:1512.03012}, 2015.

\bibitem[Dwivedi et~al.(2022)Dwivedi, Ramp{\'a}{\v{s}}ek, Galkin, Parviz, Wolf, Luu, and Beaini]{dwivedi2022long}
Vijay~Prakash Dwivedi, Ladislav Ramp{\'a}{\v{s}}ek, Michael Galkin, Ali Parviz, Guy Wolf, Anh~Tuan Luu, and Dominique Beaini.
\newblock Long range graph benchmark.
\newblock \emph{Advances in Neural Information Processing Systems}, 35:\penalty0 22326--22340, 2022.

\bibitem[Flinth and Ohlsson(2023)]{flinth2023optimization}
Axel Flinth and Fredrik Ohlsson.
\newblock Optimization dynamics of equivariant and augmented neural networks, 2023.

\bibitem[Geiger and Smidt(2022)]{geiger2022e3nn}
Mario Geiger and Tess Smidt.
\newblock e3nn: Euclidean neural networks.
\newblock \emph{arXiv preprint arXiv:2207.09453}, 2022.

\bibitem[Jain et~al.(2021)Jain, Wu, Wright, Mirhoseini, Gonzalez, and Stoica]{jain2021representing}
Paras Jain, Zhanghao Wu, Matthew~A. Wright, Azalia Mirhoseini, Joseph~E. Gonzalez, and Ion Stoica.
\newblock Representing long-range context for graph neural networks with global attention.
\newblock In A.~Beygelzimer, Y.~Dauphin, P.~Liang, and J.~Wortman Vaughan, editors, \emph{Advances in Neural Information Processing Systems}, 2021.

\bibitem[Jumper et~al.(2021)Jumper, Evans, Pritzel, Green, Figurnov, Ronneberger, Tunyasuvunakool, Bates, {\v{Z}}{\'\i}dek, Potapenko, et~al.]{jumper2021highly}
John Jumper, Richard Evans, Alexander Pritzel, Tim Green, Michael Figurnov, Olaf Ronneberger, Kathryn Tunyasuvunakool, Russ Bates, Augustin {\v{Z}}{\'\i}dek, Anna Potapenko, et~al.
\newblock Highly accurate protein structure prediction with alphafold.
\newblock \emph{Nature}, 596\penalty0 (7873):\penalty0 583--589, 2021.

\bibitem[Karhadkar et~al.(2023)Karhadkar, Banerjee, and Montufar]{karhadkar2023fosr}
Kedar Karhadkar, Pradeep~Kr. Banerjee, and Guido Montufar.
\newblock Fo{SR}: First-order spectral rewiring for addressing oversquashing in {GNN}s.
\newblock In \emph{International Conference on Learning Representations}, 2023.

\bibitem[Levi et~al.(2019)Levi, Allen, Raichoor, Baltay, BenZvi, Beutler, Bolton, Castander, Chuang, Cooper, et~al.]{levi2019dark}
Michael~E Levi, Lori~E Allen, Anand Raichoor, Charles Baltay, Segev BenZvi, Florian Beutler, Adam Bolton, Francisco~J Castander, Chia-Hsun Chuang, Andrew Cooper, et~al.
\newblock The dark energy spectroscopic instrument (desi).
\newblock \emph{The Bulletin of the American Astronomical Society}, 57\penalty0 (7), 2019.

\bibitem[Li et~al.(2018)Li, Han, and Wu]{Li_Han_Wu_2018}
Qimai Li, Zhichao Han, and Xiao-ming Wu.
\newblock Deeper insights into graph convolutional networks for semi-supervised learning.
\newblock \emph{Proceedings of the AAAI Conference on Artificial Intelligence}, 32\penalty0 (1), Apr. 2018.

\bibitem[Loshchilov and Hutter(2017)]{loshchilov2017decoupled}
Ilya Loshchilov and Frank Hutter.
\newblock Decoupled weight decay regularization.
\newblock \emph{arXiv preprint arXiv:1711.05101}, 2017.

\bibitem[Makinen et~al.(2022)Makinen, Charnock, Lemos, Porqueres, Heavens, and Wandelt]{makinen2022cosmic}
T~Lucas Makinen, Tom Charnock, Pablo Lemos, Natalia Porqueres, Alan Heavens, and Benjamin~D Wandelt.
\newblock The cosmic graph: Optimal information extraction from large-scale structure using catalogues.
\newblock \emph{arXiv preprint arXiv:2207.05202}, 2022.

\bibitem[Qi et~al.(2017)Qi, Su, Mo, and Guibas]{qi2017pointnet}
Charles~R Qi, Hao Su, Kaichun Mo, and Leonidas~J Guibas.
\newblock Pointnet: Deep learning on point sets for 3d classification and segmentation.
\newblock In \emph{Proceedings of the IEEE conference on computer vision and pattern recognition}, pages 652--660, 2017.

\bibitem[Qian et~al.(2022)Qian, Li, Peng, Mai, Hammoud, Elhoseiny, and Ghanem]{qian2022pointnext}
Guocheng Qian, Yuchen Li, Houwen Peng, Jinjie Mai, Hasan Hammoud, Mohamed Elhoseiny, and Bernard Ghanem.
\newblock Pointnext: Revisiting pointnet++ with improved training and scaling strategies.
\newblock \emph{Advances in Neural Information Processing Systems}, 35:\penalty0 23192--23204, 2022.

\bibitem[Satorras et~al.(2021)Satorras, Hoogeboom, and Welling]{satorras2021n}
V{\i}ctor~Garcia Satorras, Emiel Hoogeboom, and Max Welling.
\newblock E (n) equivariant graph neural networks.
\newblock In \emph{International conference on machine learning}, pages 9323--9332. PMLR, 2021.

\bibitem[T{\"o}nshoff et~al.(2023)T{\"o}nshoff, Ritzert, Rosenbluth, and Grohe]{tonshoff2023did}
Jan T{\"o}nshoff, Martin Ritzert, Eran Rosenbluth, and Martin Grohe.
\newblock Where did the gap go? reassessing the long-range graph benchmark.
\newblock \emph{arXiv preprint arXiv:2309.00367}, 2023.

\bibitem[Topping et~al.(2022)Topping, Giovanni, Chamberlain, Dong, and Bronstein]{topping_understanding_2022}
Jake Topping, Francesco~Di Giovanni, Benjamin~Paul Chamberlain, Xiaowen Dong, and Michael~M. Bronstein.
\newblock Understanding over-squashing and bottlenecks on graphs via curvature.
\newblock In \emph{International Conference on Learning Representations}, 2022.

\bibitem[Toshev et~al.(2024)Toshev, Galletti, Fritz, Adami, and Adams]{toshev2024lagrangebench}
Artur Toshev, Gianluca Galletti, Fabian Fritz, Stefan Adami, and Nikolaus Adams.
\newblock Lagrangebench: A lagrangian fluid mechanics benchmarking suite.
\newblock \emph{Advances in Neural Information Processing Systems}, 36, 2024.

\bibitem[Villaescusa-Navarro et~al.(2020)Villaescusa-Navarro, Hahn, Massara, Banerjee, Delgado, Ramanah, Charnock, Giusarma, Li, Allys, et~al.]{villaescusa2020quijote}
Francisco Villaescusa-Navarro, ChangHoon Hahn, Elena Massara, Arka Banerjee, Ana~Maria Delgado, Doogesh~Kodi Ramanah, Tom Charnock, Elena Giusarma, Yin Li, Erwan Allys, et~al.
\newblock The quijote simulations.
\newblock \emph{The Astrophysical Journal Supplement Series}, 250\penalty0 (1):\penalty0 2, 2020.

\bibitem[Villanueva-Domingo and Villaescusa-Navarro(2022)]{villanueva2022learning}
Pablo Villanueva-Domingo and Francisco Villaescusa-Navarro.
\newblock Learning cosmology and clustering with cosmic graphs.
\newblock \emph{The Astrophysical Journal}, 937\penalty0 (2):\penalty0 115, 2022.

\bibitem[Wang et~al.(2023)Wang, Elhag, Jaitly, Susskind, and Bautista]{wang2023generating}
Yuyang Wang, Ahmed~A Elhag, Navdeep Jaitly, Joshua~M Susskind, and Miguel~Angel Bautista.
\newblock Generating molecular conformer fields.
\newblock \emph{arXiv preprint arXiv:2311.17932}, 2023.

\bibitem[Wu et~al.(2015)Wu, Song, Khosla, Yu, Zhang, Tang, and Xiao]{Zhirong15CVPR}
Z.~Wu, S.~Song, A.~Khosla, F.~Yu, L.~Zhang, X.~Tang, and J.~Xiao.
\newblock 3d shapenets: A deep representation for volumetric shapes.
\newblock In \emph{Computer Vision and Pattern Recognition}, 2015.

\end{thebibliography}

\newpage
\appendix

\section{Dataset description}
\label{app:dataset}

\subsection{Details of simulation}\label{app:simulation}
We use the Big Sobol Sequence (BSQ) of the \emph{Quijote} simulations \citep{villaescusa2020quijote}, a collection of 32,768 $N$-body simulations designed for machine learning applications. Each simulation models the evolution of the large-scale structure of the Universe by following the dynamics of $512^3$ cold dark matter particles in a cubic comoving volume of side $\sim 1$ Gigaparsec from redshift $z=127$ to $z=0$ (present time). Dark matter halos, which are gravitationally bound structures that host galaxies, are identified in the simulations using the \emph{Rockstar} halo finder \citep{Behroozi_2012}.

The simulations are performed using the TreePM \emph{Gadget-III} code, which efficiently computes gravitational forces using a combination of a short-range tree method and a long-range particle mesh method. Each of these simulations has a different initial random seed and a value of the cosmological parameters arranged in a Sobol sequence with boundaries

\begin{align}
       \Omega_{\rm m} &\in [0.10 ; 0.50] \\
       \Omega_{\rm b} &\in [0.02 ; 0.08] \\
       h &\in [0.50 ; 0.90] \\
       n_s &\in [0.80 ; 1.20] \\
       \sigma_8 &\in [0.60 ; 1.00]
\end{align}

The initial conditions were generated at $z=127$ using 2LPT, and the simulations have been run using Gadget-III.

\subsection{Data access}\label{app:data_access}

\begin{lstlisting}[language=Python, upquote=true]
from benchmarks.galaxies.dataset import get_halo_dataset

features = ['x', 'y', 'z', 'v_x', 'v_y', 'v_z', 'M200c']  
params = ['Omega_m', 'sigma_8']

dataset, num_total = get_halo_dataset(batch_size=32 num_samples=2048, split='train', standardize=True, return_mean_std=False, seed=42, features=features, params=params, include_tpcf=True)

iterator = iter(dataset)
for _ in range(num_total // batch_size):
    x, params, tpcf = next(iterator)  # Load a batch of data

print(x.shape, params.shape, tpcf.shape)
>> (TensorShape([32, 5000, 7]), TensorShape([32, 2]), TensorShape([32, 24]))
\end{lstlisting}

The \texttt{get\_halo\_dataset} function loads the dataset with the specified batch size, number of samples, data split, and a list of desired features and cosmological parameters. The loaded data can be easily iterated over in batches, with each batch containing the point cloud features (spatial coordinates, velocities, and halo masses in this case) and the corresponding cosmological parameters ($\Omega_m$ and $\sigma_8$, in this case). There is also an option to include the pre-computed 2PCF vectors as a third output.

\section{Details of neural network architectures}
\label{app:architectures}

Below, we describe the differences between the message-passing functions of all graph neural network models used in our study. The training hyperparameters are provided in Tab. \ref{tab:models_params}. In all equations below, we use the following notation to denote the relative distance vectors and their projections onto a basis of radial Bessel functions of order $n=64$ and radial cutoff $c=0.6$:
\begin{align}
    \vec{r}_{ij}^l &= \vec{x}_i^l - \vec{x}_j^l\\
    R_{ij}^l &= B_n(\|{\vec{r}_{ij}}^l\|^2, c).
\end{align}
The Bessel basis functions are defined as $B_n(r, c) = \sqrt{\frac{2}{c}} \frac{\sin(\frac{n \pi}{c} r)}{r}$ for $r \neq 0$, and $B_n(0, c) = \sqrt{\frac{2}{c}} \frac{n \pi}{c}$.

\paragraph{MLP on 2PCF} When using the 2-point correlation function summary instead of the full point cloud, an MLP with 3 hidden layers of dimension 128 and GELU activations was used on the 24-dimensional 2PCF vectors. 

\paragraph{GNN} Message-passing GNNs consist of the following edge and node update functions in one message-passing layer,

\begin{align}
\vec{e}_{ij}^{l+1} &= \phi_e^l\left(\vec{h}_i^l, \vec{h}_j^l, \vec{e}_{ij}^l\right)\\
\vec{h}_i^{l+1} &= \phi_h^l\left(\vec{h}_i^l, \square_{j \in \mathcal{N}(i)} \vec{e}_{ij}^{l+1}\right) 
\end{align}
where all input vectors are concatenated before being fed into 3-layer MLPs $\phi_e^l$ and $\phi_h^l$. Additionally, $\square$ denotes a permutation-invariant message-passing aggregation function over the neighboring edges $\mathcal{N}(i)$ of node $i$. We select $\square$ to be defined as the mean for all of our models.

\paragraph{EGNN} The $E(n)$ Equivariant Graph Neural Network \citep{satorras2021n} edge, position, and node representation update functions are defined as 
\begin{align} \vec{e}_{ij}^{l+1} & =\phi_e^{l}\left(\vec{h}_i^l, \vec{h}_j^l,R_{ij}^{l}\right) \\ \vec{x}_i^{l+1} & =\vec{x}_i^l+C \sum_{j \neq i} \vec{r}_{ij}^l \cdot \phi_x^l\left(\vec{e}_{ij}^{l+1}\right) \\ \vec{h}_i^{l+1} & =\phi_h^l\left(\vec{h}_i^l, \square_{j \in \mathcal{N}(i)} \vec{e}_{ij}^{l+1}\right)
\end{align}

The edge update operation is invariant, depending only on the absolute distances. The node position updates are equivariant, depending linearly on the relative position vectors, gated with a nonlinear function of the invariant edge features. 

\paragraph{SEGNN} Steerable $E(3)$ Equivariant Graph Neural Networks \citep{brandstetter2021geometric} extend the EGNN by implementing the node and edge update functions as O(3) steerable MLPs $\phi$, consisting of steerable linear layers conditioned on a steerable feature $\tilde{\vec{a}} \in V_0 \oplus \ldots \oplus V_{\ell_\mathrm{max}}$ (e.g., positions and/or velocities),
\begin{equation}
    \sigma\left(W_{\tilde{\vec{a}}} \tilde{h}^l\right) := \sigma\left(\tilde{h}^l \otimes_{c g}^{W} \tilde{\vec{a}}\right) 
\end{equation}
where $\sigma$ is a gated non-linearity, $W_{\tilde{\vec{a}}}$ is a linear transformation matrix conditioned on ${\tilde{\vec{a}}}$, and $\otimes_{c g}^{W}$ is the Clebsch-Gordan tensor product that is parametrized by a collection of weights. Thus, the steerable edge and node feature updates are given by
\begin{align} \tilde{\vec{e}_{ij}}^{l+1} & =\phi_e^{l}\left(\tilde{\vec{h}_i}^l,\tilde{\vec{h}_j}^l,R_{ij}^l\right) \\ 
\tilde{\vec{h}_i}^{l+1} & =\phi_h^l\left(\tilde{\vec{h}_i}^l, \square_{j \in \mathcal{N}(i)} \tilde{\vec{e}_{ij}}^{l+1}, \tilde{\vec{a}}_i\right)
\end{align}
while the node position updates remain unchanged from eq. (11) above. In our experiments, $\tilde{\vec{a}}_i$ is optionally defined via the node velocities.

The steerable MLPs allow the network to leverage richer geometric information and express anisotropic interactions. The updates are conditioned on steerable node and edge \emph{attributes}, which can inject additional physical information about the local environment into the updates while maintaining end-to-end equivariance.

\paragraph{NequIP} The Neural Equivariant Interatomic Potential  architecture \citep{batzner20223} also utilizes steerable features in the message-passing updates. In the edge updates, NequIP applies a linear transformation to the incoming node features and computes the spherical harmonic projections of the normalized relative position vectors, which are combined using a tensor product. This is modulated by a nonlinear radial function implemented as an MLP acting on the relative distances. The node updates combine the aggregated messages with the previous node features using a gated nonlinearity. 

The edge and node updates are thus defined as

\begin{align}
    a_{ij}^l &= \operatorname{Norm}\left(Y_m^{(\ell)}\left(\vec{r}_{ij}^l\right)\right) \\
    \vec{e}_{ij}^{l+1} &= \phi_e^l(R_{ij}^{l})  \cdot [W_{i}^l \vec{h}_i^l, a_{ij}^l] \otimes a_{ij}^l\\
    \vec{h}_i^{l+1} &= \phi_h^l \left( \frac{\square_{j \in \mathcal{N}(i)} \vec{e}_{ij}^{l+1}}{\sqrt{|E|}} \right). \\
\end{align}
where the spherical harmonics are normalized via the integral norm such that $\int_{S^2} Y_m^{\ell}(x)^2 d x=1$.

While both are $E(3)$ equivariant, SEGNN uses a more expressive steerable MLP conditioned on the spherical harmonic embedding of the relative position vectors, while NequIP uses a simpler nonlinear radial function to gate linear and spherical harmonic projections of the input features. These choices reflect a trade-off between expressiveness and computational efficiency, with NequIP prioritizing the latter, tailored for its original purpose of efficiently learning interatomic potentials where angular and radial dependencies are crucial and separable.

\paragraph{PointNet++} PointNet++ \citep{qian2022pointnext} is an extension of the original PointNet architecture \citep{qi2017pointnet} designed to capture both local and global structures in point clouds. Unlike the original PointNet, which treats each point independently, PointNet++ introduces a hierarchical learning framework. It recursively applies PointNet at multiple scales, progressively downsampling the point cloud and learning features at different levels of granularity. This enables the network to capture fine-grained local features as well as broader contextual information. 

The PointNet++ architecture employs farthest point sampling (FPS) to select a subset of representative points and groups neighboring points within a defined radius for local feature extraction. These hierarchical groupings are followed by feature pooling and graph-based operations to coarsen the point cloud representation. The combination of these steps allows PointNet++ to learn spatially aware representations that are crucial for processing 3D point clouds.

In each downsampling layer \( i \in \{1, \dots, \texttt{n\_downsamples}\} \), the number of nodes is reduced from \( \texttt{n\_nodes} \) by dividing the original node set by a predefined downsampling factor, \texttt{d\_downsampling\_factor}.
At each layer, a Graph Neural Network (GNN) is applied to the graph to obtain updated node embeddings \( z \), such that
\begin{align}
z_i^{l+1} = \text{GNN}(\vec{h}_i^{l}, \vec{x}_i^{l}, \{\vec{e}_{ij}^{l}\}_{j\in \mathcal{N}(i)}),
\end{align}
where \( z_i^{l+1} \in \mathbb{R}^{n_{\text{nodes}} \times d} \) represents the new node embeddings at this layer. After applying the GNN, we perform a \textit{sample and group} operation to downsample the set of nodes and create a hierarchical representation. This operation consists of two steps: sampling representative points (centroids) and grouping the remaining points around these centroids. First, the centroids are selected from the set of node positions \( \mathbf{X} \in \mathbb{R}^{n_{\text{nodes}} \times d} \). The sampling is performed using Farthest Point Sampling (FPS), which selects \( n_{\text{centroids}} \) representative points:
\begin{align}
\mathbf{X}_{\text{centroids}} = \text{FPS}(\mathbf{X}, n_{\text{centroids}}),
\end{align}
where \( \mathbf{x}_{\text{centroids}} \in \mathbb{R}^{n_{\text{centroids}} \times d} \) represents the set of centroids chosen from the original node positions \( \mathbf{x} \).

After selecting the centroids, each point \( \mathbf{x}_i \) from the original set is grouped with the nearest centroid. 
The distance matrix \( \mathbf{D} \in \mathbb{R}^{n_{\text{nodes}} \times n_{\text{centroids}}} \) is then transformed into an assignment matrix \( \mathbf{S} \in \mathbb{R}^{n_{\text{nodes}} \times n_{\text{centroids}}} \) using a row-wise softmax:
\begin{align}
S_{ij} = \frac{\exp(-d_{ij})}{\sum_{k=1}^{n_{\text{centroids}}} \exp(-d_{ik})}.
\end{align}

The assignment matrix \( S \) represents the association between nodes and centroids, where each entry \( S_{ij} \) indicates the probability that node \( i \) is assigned to centroid \( j \). The matrix \( S \) is subsequently used to pool features, coarsen the graph, or aggregate information for hierarchical graph processing.

\newpage

\section{Model hyperparameters}\label{app:hyperparams}

Table \ref{tab:models_params} contains the hyperparameter values used for each experiment across all models, which we refer to by their corresponding names in the \texttt{eqnn-jax} code repository at \url{https://github.com/smsharma/eqnn-jax}. For the remaining hyperparameters, we conducted an extensive sweep using a grid search method. The hyperparameter ranges are outlined in Table \ref{tab:sweeps}.

\begin{table*}[!htbp]
\centering
\setlength{\tabcolsep}{4pt}
\renewcommand{\arraystretch}{1.1}
\begin{tabular*}{0.7\textwidth}{@{\extracolsep{\fill}} l c c c c c @{}}
\toprule
\textbf{Hyperparameter} & \textbf{Value}\\ 
\midrule
\texttt{d\_hidden} & 128 \\ 
\texttt{n\_layers} & 3  \\ 
\texttt{mlp\_readout\_widths} & (4, 2, 2)\\ 
\texttt{residual} & True \\ 
\texttt{scalar\_activation} & gelu \\ 
\texttt{gate\_activation} & sigmoid \\ 
\texttt{spherical\_harmonic\_norm} & integral\\ 
\bottomrule
\end{tabular*}
\caption{Shared hyperparameters for each model (where relavant) on all tasks.}
\label{tab:models_params}
\end{table*}

\begin{table*}[!htbp]
\centering
\setlength{\tabcolsep}{4pt}
\renewcommand{\arraystretch}{1.1}
\begin{tabular*}{0.7\textwidth}{@{\extracolsep{\fill}} l c @{}}
\toprule
\textbf{Hyperparameter} & \textbf{Values} \\ 
\midrule
\texttt{learning\_rate} & \{0.001, 0.0001, 0.00001\} \\ 
\texttt{decay} & \{0.0001, 0.00001\} \\ 
\texttt{n\_radial\_basis} & \{32, 64\} \\ 
\texttt{k} & \{5, 10\} \\ 
\texttt{message\_passing\_agg} & \{Sum, Mean, Max\} \\ 
\texttt{readout\_agg} & \{Sum, Mean, Max\} \\ 
\texttt{message\_passing\_steps} & \{2, 4, 6\} \\ 
\midrule
\texttt{n\_downsamples} & \{2, 3, 4\} \\ 
\texttt{d\_downsampling\_factor} & \{2, 5, 10\} \\ 
\texttt{k\_downsample} & \{10, 20\} \\ 
\texttt{r\_downsample} & \{0.05, 0.2\} \\ 
\texttt{combine\_hierarchies\_method} & \{Mean, Concat\} \\ 
\bottomrule
\end{tabular*}
\caption{Hyperparameter ranges used for grid search. The parameters below the midrule were only use for the PointNet++ architecture.}
\label{tab:sweeps}
\end{table*}

\section{Ablation study of attention-based aggregation}\label{app:gnn_attn}

We perform an ablation study where we replace the standard aggregation mechanism within the GNN layers with attention-based local and global aggregation. In particular, we modify the GNN layers to incorporate multi-head self-attention mechanisms in place of traditional neighborhood aggregation. Rather than relying solely on proximity-based neighbors, the edge weights are modulated by an attention score computed from other message passing components. These weights dynamically adjust the importance of both local and distant nodes, allowing the model to better capture complex relationships across the graph. 

The results demonstrate that incorporating attention-based mechanisms into the readout layer of the GNN significantly improves performance. However, the model that combines both local and global attention does not show the same improvement and even results in increased error. Similarly, the invariant attention model performs worse than both the standard and global attention-based models. 

\begin{table*}[!htbp]
\centering
\setlength{\tabcolsep}{4pt}
\renewcommand{\arraystretch}{1.2}
\begin{tabular*}{\textwidth}{@{\extracolsep{\fill}} l T{1.75cm} T{2cm} T{1cm}@{}}
\toprule
\multicolumn{1}{c}{Model} & \multicolumn{1}{c}{$\Omega_m$} & \multicolumn{1}{c}{$\sigma_8$} & \multicolumn{1}{c}{Params.} \\ 
\midrule
GNN (mean agg.) & $2.77 \pm 0.41$  & $4.84 \pm 2.90$ & 1441k \\ 
GNN (global attn. agg.) & $\mathbf{2.60 \pm 0.06}$  & $\mathbf{2.84 \pm 0.19}$ & 915k \\ 
GNN (local + global attn. agg.) & $3.00 \pm 0.14$  & $8.82 \pm 3.15$ & 913k \\ 
GNN (invariant attn.) & $3.62 \pm 0.01$  & $13.33 \pm 0.01$ & 967k \\ 
\bottomrule
\end{tabular*}
\caption{Comparison of different types of local and global aggregation on the graph-level tasks. The best results for each task are shown in \textbf{bold}.}
\label{tab:gnn_attn}
\end{table*}

\end{document}